\documentclass{article}

\usepackage[english]{babel}

\usepackage[letterpaper,top=1in,bottom=1in,left=1in,right=1in,marginparwidth=1.75cm]{geometry}

\usepackage{amsmath}
\usepackage{listings}
\usepackage{graphicx}
\usepackage[colorlinks=true, allcolors=blue]{hyperref}
\usepackage{amsfonts,amssymb}
\usepackage{indentfirst}
\usepackage{float}
\title{A new approach for image segmentation based on diffeomorphic registration and gradient fields}
\author{Junchao Zhou}
\date{Faculty advisor: Nicolas Charon}

\begin{document}
\maketitle


\section{Introduction}

\subsection{Image segmentation}
    
    Image segmentation is one of the core problem of image processing and computer vision. It generally consists in the extraction of the boundaries of predominant objects in a 2D or 3D image. This usually allows to simplify and/or change the representation of an image into a model more meaningful and easier to analyze \cite{cvbook, pips}. There are many different methods that have been proposed to address the problem of image segmentation which can be broadly divided into the following categories. 
    
    \paragraph{Edge Detection Approach} Edge detection is a relatively well-developed field and an approach used frequently for segmenting images based on abrupt (local) changes in intensity \cite{dip}. The edge-detection methods are based on filtering an image with one or more kernels, and factors such as image noise and the nature of edges themselves can be also taken into account. Two classical examples are the Marr-Hildreth edge detector and the Canny edge detector. However, a downside of those methods is the fact that they usually require a post-processing step to reconstruct a single boundary curve or surface from the detected edges. 
        
    \paragraph{Variational Methods} Variational methods (or active contour models), on the other hand, operate by directly optimizing the contour curve or surface to align it with the most salient object boundaries in the image. They rely on the design and minimization of a certain energy functional which usually consists in the combination of a data fitting term and a regularizing terms. Well-known models of this kind include for example the Mumford-Shah functional \cite{mumford1989optimal}, the Chan-Vese approach \cite{chan2001active} as well as the class of snakes algorithms \cite{kass1988snakes}.
        
    \paragraph{Deep Learning Methods} More recently, the development of deep learning has lead to new numerical methods with high performance for image segmentation \cite{dlpaper}. These methods generally train a neural network in a supervised or unsupervised fashion to extract boundaries in an image. One of the difficulty of deep learning approaches, however, is that they typically require large amount of training data and do not generalize well for images of a different type than those of the training set. This can pose some issues in e.g. certain areas of medical imaging where the number of data samples may be relatively small.    

\subsection{Overview and motivation of our approach}
    The approach we investigated in this project falls within the previous category of variational methods for image segmentation and we specifically focused on 2D images for simplicity. As with usual variational approaches, we formulate the problem as the minimization of a functional over an evolving curve. However, our model transforms a template curve by a diffeomorphic transformation of the 2D space for which we rely on the framework of large deformation diffeomorphic metric mapping (LDDMM) \cite{beg2005computing} in order to align the curve with the dominant gradient directions in the image. To achieve this, we further introduce a particular loss function between curves and gradient fields which is derived from the varifold representation of geometric objects \cite{charon2017framework}. 
    
    The implementation of our approach is done in Python and uses the PyKeops library for efficient GPU computations and optimization.

\section{Images and gradient fields}
\subsection{Images}
    Our goal is to segment images from different modalities, such as binary images of objects, medical images (e.g. CT scans or MRI), and in a way that is robust to imperfections like noise... From the continuous viewpoint, a 2D image is a function $I:\Omega \rightarrow \mathbb{R}$ where $\Omega\subset \mathbb{R}^2$ is the domain of $I$. From the discrete perspective, an image is represented by a grid of pixels where each pixel can be viewed as a sample of the continuous image. Thus, denoting $p_i=(x_i,y_i) \in \mathbb{R}^2$ the integer 2D coordinates of the i-th pixel of $I$, we shall write $I(p_i)$ for the value of the image at this pixel.

    
    

\subsection{Gradient Field}
\begin{figure}[H]
        \centering
	    \begin{tabular}{ccc}
	    \rotatebox{90}{\phantom{aaaaaa}Bottle Image}
	    \includegraphics[height=5cm, width=5cm]{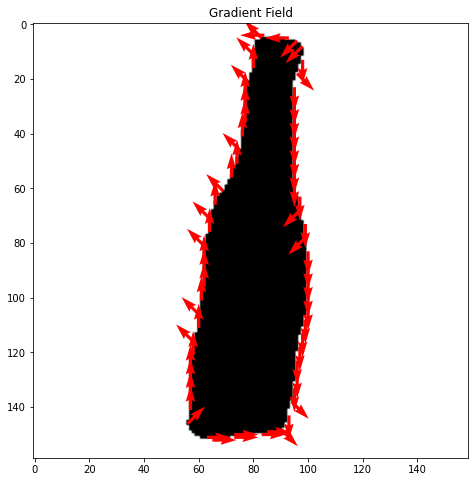}
	    \end{tabular}
	    \caption{Gradient field.} 
	    \label{fig:gradient_field}
    \end{figure}

    For the purpose of segmentation, we will need to first extract the gradient field from the image, and filter out the least significant gradient vectors based on a threshold on the norm of these gradients. 
    The gradient $\nabla I$ at each pixel $p_i=(x_i,y_i)$ is given by $\nabla I = (\partial_x I(p_i), \partial_y I(p_i))^T$ which can be approximated, in the discrete setting, by e.g. the centered finite difference scheme:
$$
\nabla I(p_i)={\begin{bmatrix}\frac{I(x_i+1,y_i)-I(x_i-1,y_i)}{2}\\\frac{I(x_i,y_i+1)-I(x_i,y_i-1)}{2}\end{bmatrix}}.
$$
Note that one can further apply a Gaussian filter with small variance to the estimated gradient field as a way to reduce the effect of noise on the direction of the vectors. An example of gradient field is shown in Figure \ref{fig:gradient_field}. 
    
\section{Model}

\subsection{Large Deformation Diffeomorphic Metric Mapping (LDDMM)}
\label{ssec:LDDMM}
    The LDDMM framework \cite{beg2005computing} consists in modelling diffeomorphic transformations of the entire 2D space by flowing time-dependent smooth velocity fields $(t,x) \in [0,1] \times \mathbb{R} ^2\mapsto v(t,x)$ that will in turn transport points of the template curve. We consider a fixed Hilbert space $V$ of smooth vector fields over $\mathbb{R}^2$ and time-dependent vector fields $v \in L^2([0,1],V)$.  
    
    Given any $v \in L^2([0,1],V)$, the flow map $(t,x) \mapsto \varphi(t,x)$ of the vector field $v$ is obtained by integration of the ODE $\partial_t\varphi(t,x) = v(t,\varphi(t,x))$ with initial conditions $\varphi(0,x)=x$, in other words it is such that for all $t \in [0,1]$ and $x \in \mathbb{R}^2$:
    $$
    \varphi(t,x) = x + \int_0^t v(s,\varphi(s,x)) ds 
    $$
    
\begin{figure}[H]
		\begin{tabular}{ccc}
		 \includegraphics[trim = 15mm 15mm 15mm 15mm,clip,height=5cm,width=5cm]{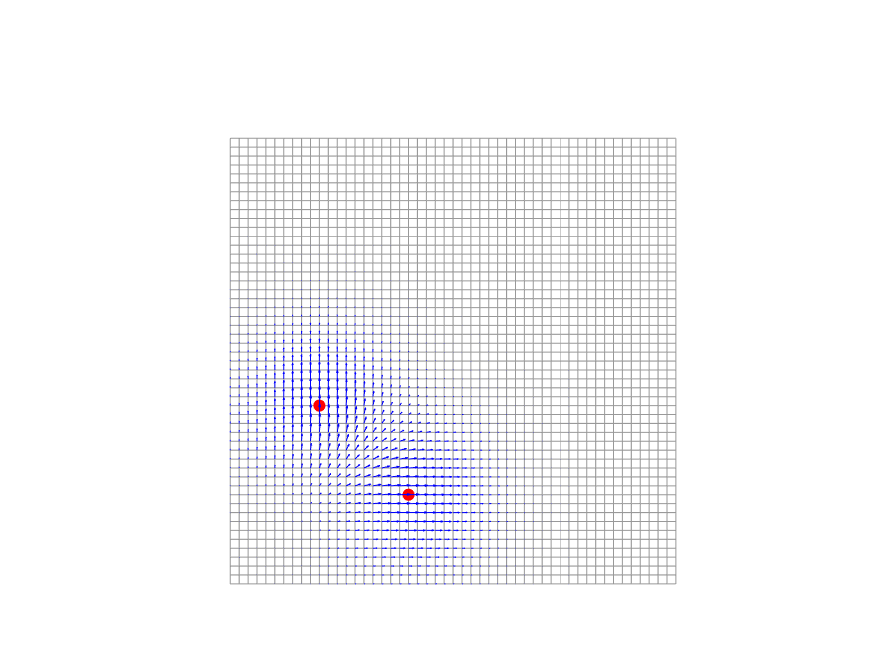}&\includegraphics[trim = 15mm 15mm 15mm 15mm,clip,height=5cm,width=5cm]{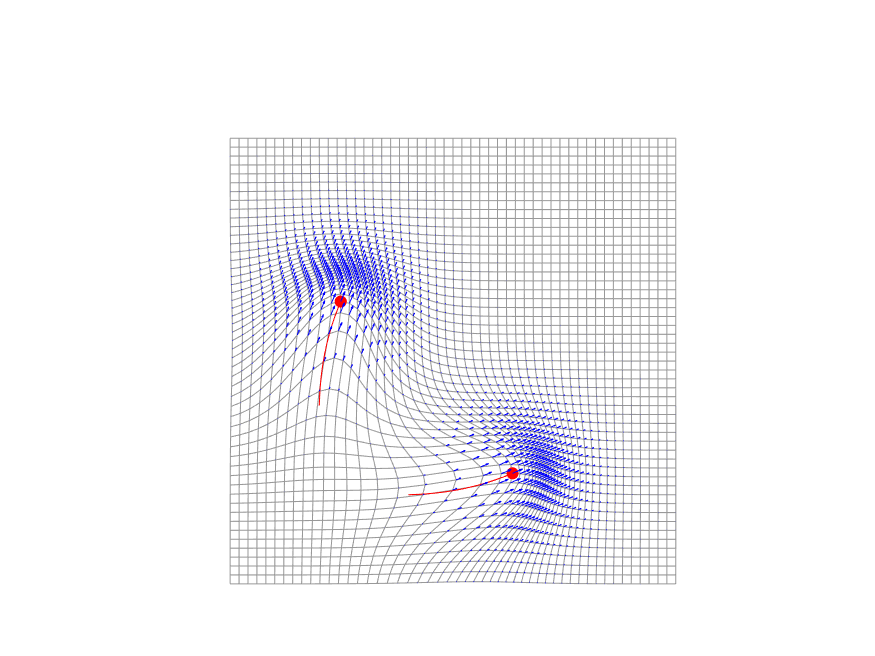}&\includegraphics[trim = 15mm 15mm 15mm 15mm,clip,height=5cm,width=5cm]{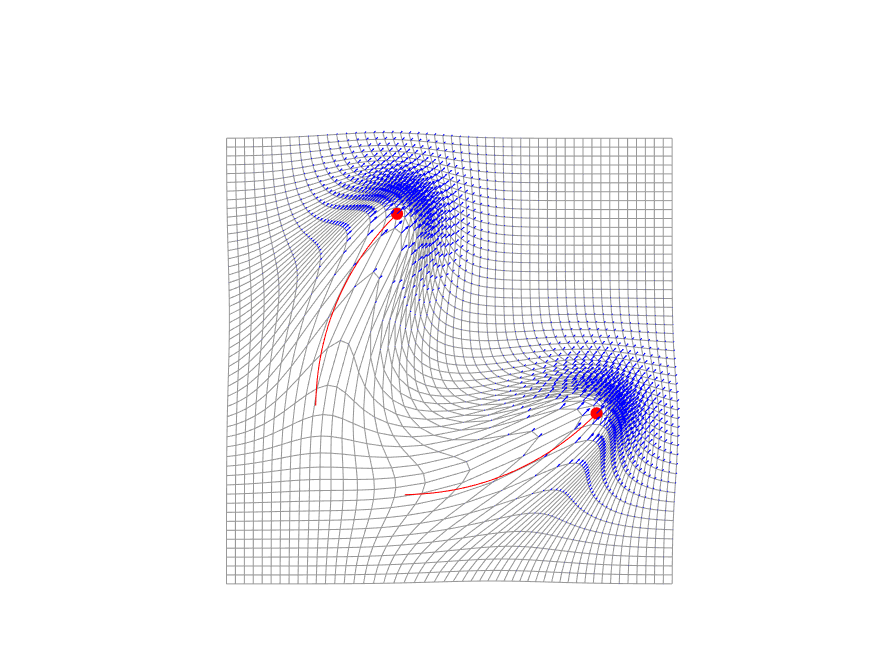}\\
		 $\varphi(0,\cdot)$ & $\varphi(1/2,\cdot)$ & $\varphi(1,\cdot)$
		\end{tabular}
		\caption{Flow map of a time-dependent vector field with the resulting deformation of the 2D space shown through the transformed grid. The trajectories along the flow of two specific points are displayed in red.} \label{fig:flow_map}
    \end{figure}

    As follows from the results of \cite{Younes2019} Chap. 7, with the appropriate conditions on the regularity of $v$, each mapping $x \mapsto \varphi(t,x)$ is a diffeomorphism of $\mathbb{R}^2$. An illustration of those maps is shown in Figure \ref{fig:flow_map}.  Moreover the set of final maps $\varphi(1,\cdot)$ for the different choices of vector fields $v$ form a group under composition and the LDDMM model provides a natural definition of metric regularization for those maps given by:
    \begin{equation*}
        \text{Reg}(v) = \int_0^1 \|v(t,\cdot)\|_V^2 dt .
    \end{equation*}
    
\subsection{Loss function}
\label{ssec:loss}
In order to enforce the alignment of the segmentation curve with the boundaries of the main objects in the image, we introduce a loss function that will measure a notion of discrepancy between the curve (source shape) and the distribution of gradients in the image (target). This can be achieved by relying on the flexibility of the class of fidelity metrics introduced for curve and surface registration in \cite{charon2017framework,charon2020fidelity}, which are based on the representation of shapes as varifolds and the construction of positive definite kernels on varifold spaces. 

In our case of interest, we may represent the image $I$ by its list of pixel positions $p_i \in \mathbb{R}^2$ together with the unit gradient $\nabla I(p_{i})/|\nabla I(p_{i})| \in \mathbb{S}^1$ of the image at pixel $p_i$, computed as explained above. Similarly, the discrete curve $c$ is represented by its list of vertices $x_i$ together with the normals $\vec{n}_i$ to the edge vectors. Both objects are then viewed as distributions of positions $\times$ directions, which corresponds to the setting of discrete varifolds. The kernel metrics considered in \cite{charon2017framework} provide a notion of Hilbert norm and inner product between those objects which takes the generic form:
\begin{equation}
\label{eq:var_inner_product}
\langle I, c \rangle_{\mathcal{V}} = \sum_{i=1}^{P} \sum_{j=1}^{N} \left|\vec{n}_{j}\right| \rho \left(\left|p_{i}-x_{j}\right|\right) \gamma\left(\frac{\nabla I(p_{i})}{\left|\nabla I(p_{i})\right|} \cdot \frac{\vec{n}_{i}}{\left|\vec{n}_{i}\right|}\right)
\end{equation}
where $\rho$ and $\gamma$ are two functions that specify the particular kernel function on the space of varifolds. In all the experiments presented below, we take a Gaussian kernel for $\rho$ i.e. $\rho(|x-y|)=e^{-|x-y|^2/\sigma^2}$ where $\sigma>0$ corresponds to a scale hyperparameter in the model, and for $\gamma$ the Cauchy-Binet kernel on the circle $\mathbb{S}^1$ i.e. $\gamma(u \cdot v) = (u \cdot v)^2 = \cos^2(\theta_{u,v})$, $\theta_{u,v}$ denoting the angle between the two unit vectors $u$ and $v$. The loss function between the curve $c$ and the image $I$ can be then taken as the squared varifold kernel norm, that is:
$$
L_0(c,I) = \|I - c \|_{\mathcal{V}}^2 = \langle c, c \rangle_{\mathcal{V}} +\langle I, I \rangle_{\mathcal{V}} -2 \langle I, c \rangle_{\mathcal{V}}.
$$

\begin{figure}
	\begin{tabular}{ccc}
	\rotatebox{90}{\phantom{aaaaaa}Child Image}
	\includegraphics[height=4.8cm, width=4.8cm]{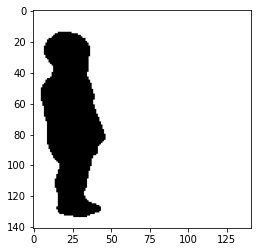}
	&\includegraphics[height=4.8cm, width=4.8cm]{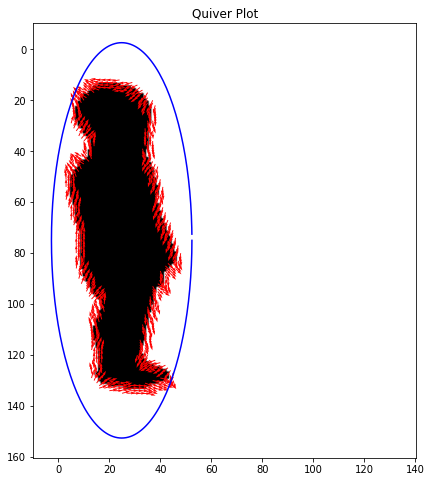}\\
	Source & Gradient Field
	\end{tabular}

\begin{tabular}{ccccc}
    \rotatebox{90}{\phantom{a}Without Weight Change}
    &\includegraphics[height=3.5cm, width=3.5cm]{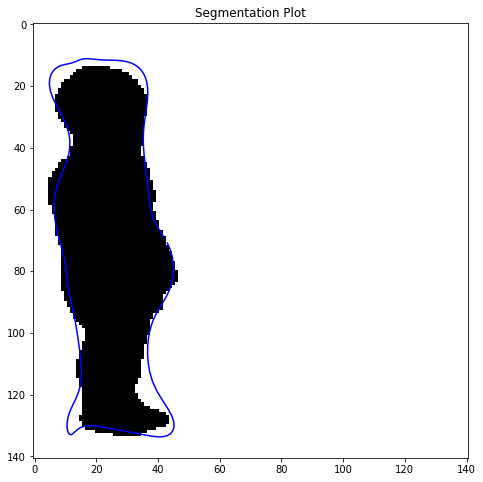}
    &\includegraphics[height=3.5cm, width=3.5cm]{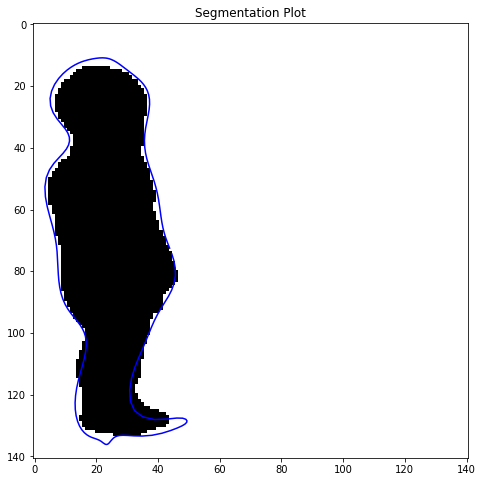}
    &\includegraphics[height=3.5cm, width=3.5cm]{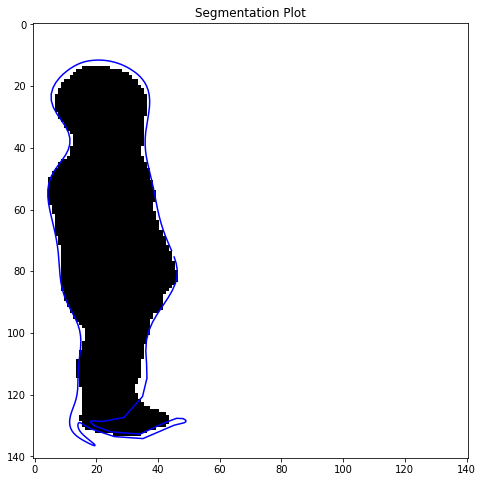}
    &\includegraphics[height=3.5cm, width=3.5cm]{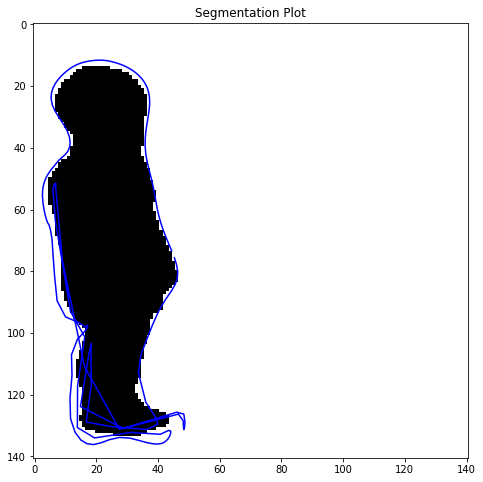}\\
    
    \rotatebox{90}{\phantom{a}Weight Change}
    &\includegraphics[height=3.5cm, width=3.5cm]{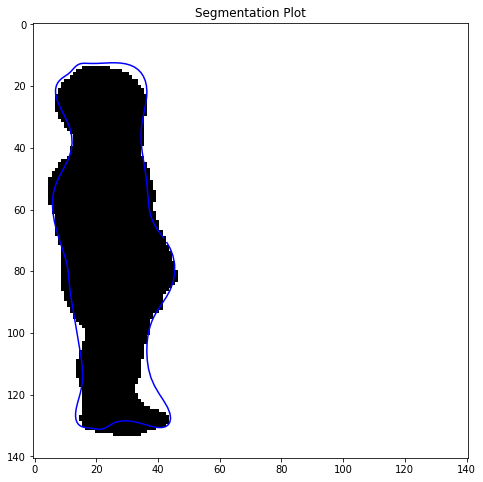}
    &\includegraphics[height=3.5cm, width=3.5cm]{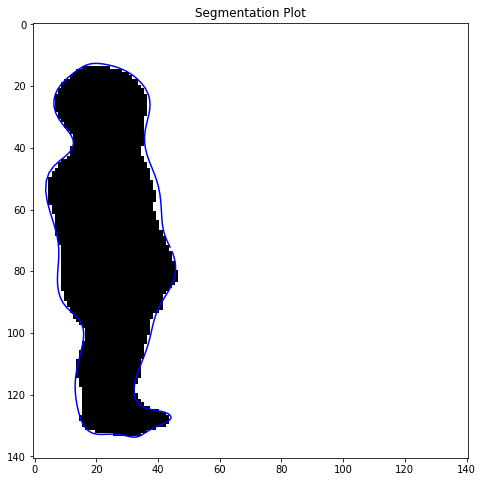}
    &\includegraphics[height=3.5cm, width=3.5cm]{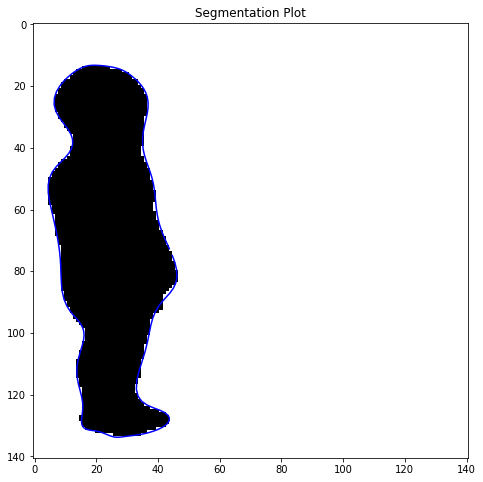}
    &\includegraphics[height=3.5cm, width=3.5cm]{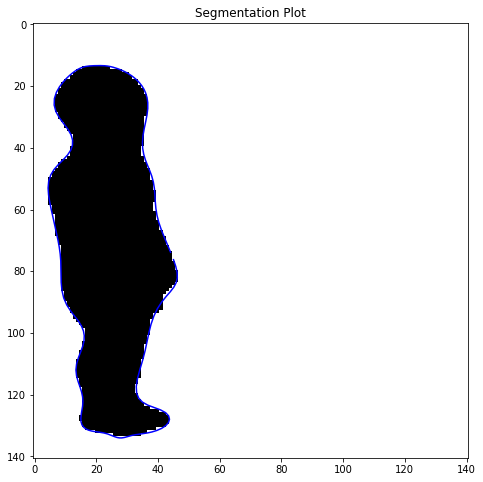}\\
    & 15 iterations & 30 iterations & 100 iterations & 300 iterations 
\end{tabular}	
\caption{Comparison of the registration results for the original varifold loss function $L_0$ (first row) vs the reweighted version $L_1$ (second row).} \label{fig:result_child}
\end{figure}

One of the remaining issues with the loss function defined above is that the choice made of setting the image gradients to unit length in the representation of the image is rather arbitrary in contrast with the curve for which the magnitude of the normal vectors $\vec{n}_i$ correspond to a real geometric feature, its local length. This can result in critical mass imbalances between the varifold associated to the source and target shapes, and lead the registration procedure to estimate unnatural deformations in an attempt to compensate for this mass discrepancy. This can be seen with the results displayed in Figure \ref{fig:result_child}.  As a way to address that issue, we can instead consider measuring the discrepancy between the image gradient field and an optimally reweighted version of the curve $c$, similar to the idea proposed in \cite{hsieh2021diffeomorphic}. This means defining the new loss function $L_1$ as:
\begin{equation*}
    L_1(c,I) = \inf_{\alpha\geq 0} \|I - \alpha \cdot c \|_{\mathcal{V}}^2
\end{equation*}
where $\alpha \cdot c$ here specifically denotes the distribution of vertices and normals given by the same vertices $x_i$ and the rescaled normals $\alpha \vec{n}_i$. It is easy to see, by expanding the squared varifold norm, that the above is a quadratic function of $\alpha$ and thus the minimum is attained for $\alpha = \langle I, c \rangle_{\mathcal{V}} / \langle c, c \rangle_{\mathcal{V}}^2$ which gives:
\begin{equation}
\label{eq:loss_weight}
    L_1(c,I) = \langle I, I \rangle_{\mathcal{V}} - \frac{\langle I, c \rangle_{\mathcal{V}}^2}{\langle c, c \rangle_{\mathcal{V}}^2}.
\end{equation}

\subsection{Optimization problem}
Eventually, under the framework of the previous sections, we may formulate the problem of segmentation as that of finding an optimal deformation of the initial template curve $c_0$ that will minimize the sum of the loss and regularization function. In other words, we look for a deformation field $v \in L^2([0,1],V)$ that minimizes the following total cost functional:
$$
E(v) = \text{Reg}(v) + L_1(I,c_1)
$$
in which $\text{Reg}(v)$ is the deformation energy of the LDDMM model introduced in Section \ref{ssec:LDDMM}, $c_1$ is the template curve $c_0$ that is deformed by the diffeomorphism $\varphi(1,\cdot)$ and $L_1(I,c_1)$ measures the discrepancy between this deformed curve point positions and normal directions with those of the image gradient fields, as defined in Section \ref{ssec:loss}. The minimization of $E$ with respect to $v$ can be thought as an optimal control problem which can be reduced to the minimization over a finite-dimensional set of variables attached to the curve vertices known as the momenta, as described in details in \cite{beg2005computing,Younes2019}. 
    
\section{Implementation in Python}

\subsection{The Pykeops library}
The implementation of both the LDDMM deformation approach as well as the loss function relies on kernel functions that need to be evaluated and differentiated many times throughout the procedure. For instance, the computation of a single varifold inner product \eqref{eq:var_inner_product} involves a quadratic number $NP$ of kernel function evaluations. This constitutes the main numerical bottleneck of the proposed segmentation model. In order to efficiently implement our approach, we make use of a particular Python library named \textit{Keops} (Kernel Operations on the GPU)\footnote{\url{https://www.kernel-operations.io/}}. The KeOps library provides fast and memory-efficient routines for kernel evaluations with associated automatic differentiation on the GPU \cite{keopspaper}, which is further compatible with the NumPy and PyTorch libraries. In addition, Keops contains a built-in implementation of the LDDMM deformation and energy for point sets. This reduces our task mainly to the coding of the new loss function which we discuss next.

\paragraph{Loss Function} The following shows the implementation of the loss function of \eqref{eq:loss_weight} in Keops for a generic kernel function $K$. It first converts the discrete curve given by the vertex list VS and edges FS into the corresponding set of points and normal vectors before evaluating the kernel loss with respect to the gradient field of $I$ which is stored by the list of pixel positions pT, gradient directions gT.  
\begin{lstlisting}[language=Python]
def lossVarifoldSurfWeight(FS, pT, gT, mT, K):
    def get_center_length_normal(F, V):
        V0, V1 = (
            V.index_select(0, F[:, 0]),
            V.index_select(0, F[:, 1]),
        )
        centers, tang = (V0 + V1) / 2, V1 - V0
        length = (tang**2).sum(dim=1)[:, None].sqrt()
        return centers, length, tang / length

    cst = (mT * K(pT, pT, gT, gT, mT)).sum()

    def loss(VS):
        CS, LS, NSn = get_center_length_normal(FS, VS)
        return (
            cst
            - ((LS * K(CS, pT, NSn, gT, mT)).sum()**2)
            /((LS * K(CS, CS, NSn, NSn, LS)).sum())
        )

    return loss
\end{lstlisting}

    
    \paragraph{Kernel function} In the experiments, we use the particular kernel function described in Section \ref{ssec:loss} for which 
    $$(K(x,y,u,v)b)_i = \sum_j \exp(-\gamma\|x_i-y_j\|^2) \langle u_i,v_j\rangle^2 b_j$$
    and its Keops implementation is shown below.
\begin{lstlisting}[language=Python]   
def GaussLinKernel(sigma):
    x, y, u, v, b = Vi(0, 2), Vj(1, 2), Vi(2, 2), Vj(3, 2), Vj(4, 1)
    gamma = 1 / (sigma * sigma)
    D2 = x.sqdist(y)
    K = (-D2 * gamma).exp() * (u * v).sum() ** 2
    return (K * b).sum_reduction(axis=1)
\end{lstlisting}
    
\subsection{Optimization}

    \paragraph{Optimization approach} As mentioned above, we need to solve an optimal control problem over the deformation of the initial curve $c_0$. We follow the same shooting strategy as in \cite{charon2017framework} or \cite{Younes2019} Chapter 10, which reduces the problem to the optimization of the initial momenta of the diffeomorphism. We use the Adam optimization algorithm \cite{adam} of the PyTorch library to minimize the resulting cost function. The gradient of the cost is computed automatically using the autodiff functionalities of Keops. The name Adam is derived from adaptive moment estimation as the optimizer uses estimations of the first and second moments of the gradient to adapt the learning rate for each variable. 
    
    \paragraph{Initialization} Regarding the choice of template curve $c_0$, we construct an ellipse as the initial outline for the part of the image to be segmented. The position and size of the ellipse is automatically chosen such that the ellipse surrounds most of the target object. The coordinate of the ellipse centre is basically generated by calculating the mean of $x$ and $y$ coordinates of the gradient field, and the width and height of the ellipse is related to the variation of the image matrix. The ellipse will gradually evolve to the contour of the target object during the optimization process.

\section{Result}

\subsection{Results on binary images}
\begin{figure}
\centering
\begin{tabular}{ccccc}
    \rotatebox{90}{\phantom{aaaaa}Bottle Image}
    &\includegraphics[height=3.5cm, width=3.5cm]{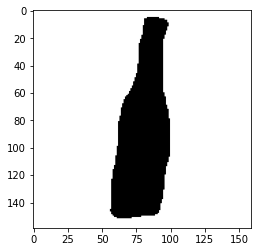}
    &\includegraphics[height=3.5cm, width=3.5cm]{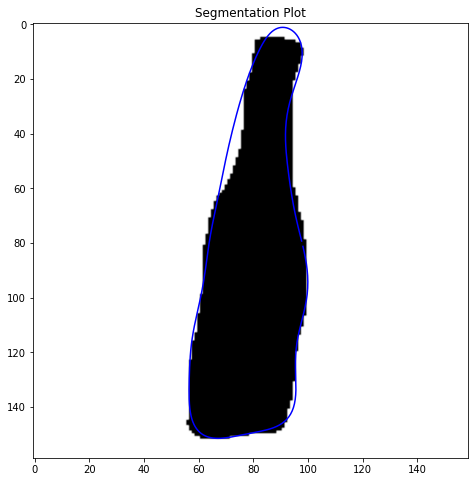}
    &\includegraphics[height=3.5cm, width=3.5cm]{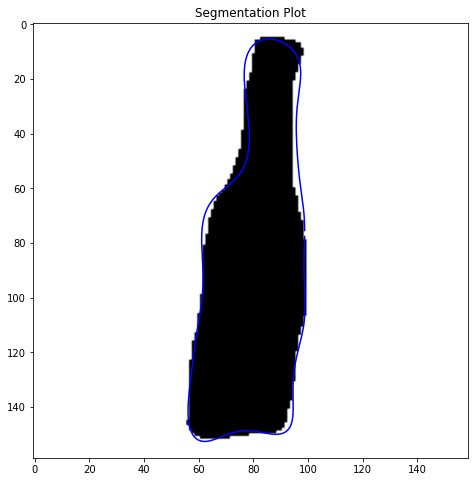}
    &\includegraphics[height=3.5cm, width=3.5cm]{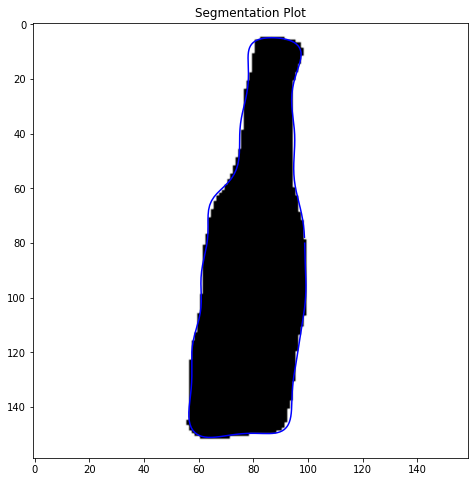}\\
    
    \rotatebox{90}{\phantom{aaaaa}Turtle Image}
    &\includegraphics[height=3.5cm, width=3.5cm]{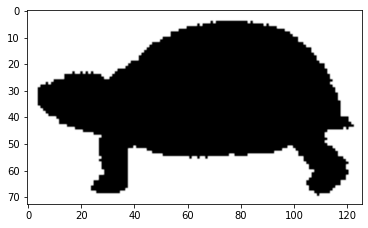}
    &\includegraphics[height=3.5cm, width=3.5cm]{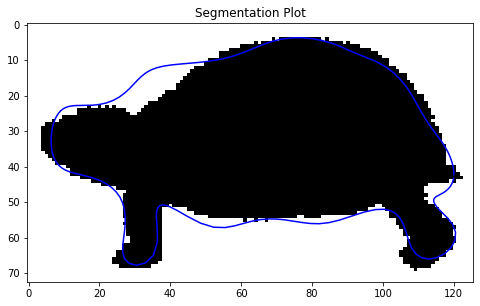}
    &\includegraphics[height=3.5cm, width=3.5cm]{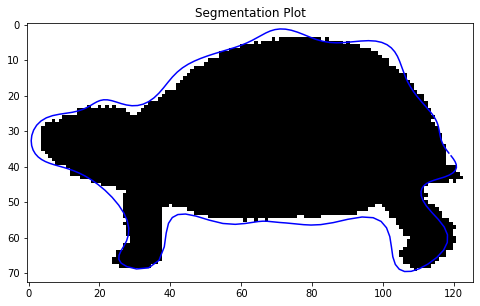}
    &\includegraphics[height=3.5cm, width=3.5cm]{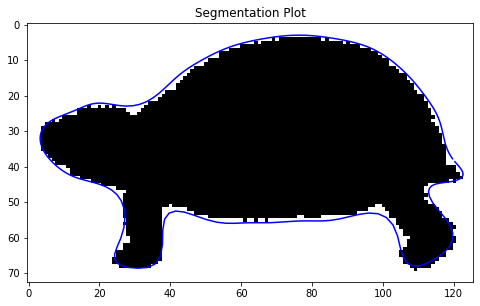}\\
    
    \rotatebox{90}{\phantom{aaaaa}Child Image}
    &\includegraphics[height=3.5cm, width=3.5cm]{figures/child_0.png}
    &\includegraphics[height=3.5cm, width=3.5cm]{figures/child_2_withweightchange_15iter.png}
    &\includegraphics[height=3.5cm, width=3.5cm]{figures/child_3_withweightchange_30iter.png}
    &\includegraphics[height=3.5cm, width=3.5cm]{figures/child_f_withweightchange_300iter.png}\\
    & Source & Processing & Processing & Final Segmentation 
\end{tabular}	
\caption{Segmentation results on binary Images from the Kimia database.} 
\label{fig:binary_image}
\end{figure}

We start with some results of our approach on simple binary images from the Kimia database \footnote{\url{https://github.com/mmssouza/kimia99}}. Figures \ref{fig:result_child} and  \ref{fig:binary_image} show the obtained deformed curves $c_1$ for three different images and throughout the iterations of the optimization algorithm.  

\subsection{Results on medical images}
\begin{figure}
\centering
	\begin{tabular}{ccc}
	\rotatebox{90}{\phantom{aaaaaaaaa}CT Scan}
	\includegraphics[height=4.8cm, width=4.8cm]{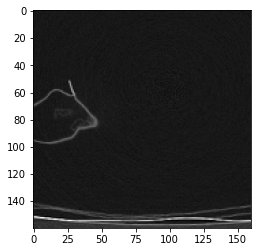}
	&\includegraphics[height=4.8cm, width=4.8cm]{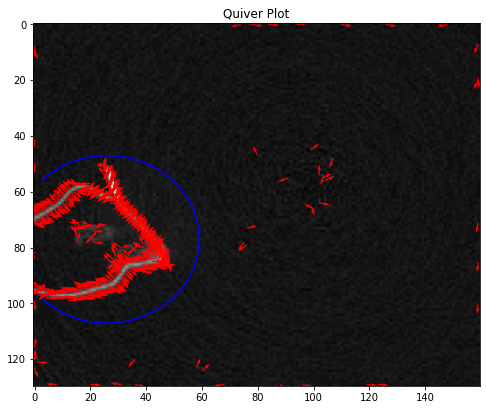}
	&\includegraphics[height=4.8cm, width=4.8cm]{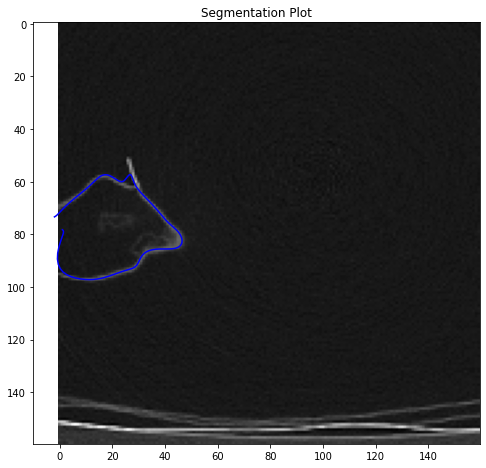}\\
	Source & Gradient Field & Segmentation
	\end{tabular}
\caption{Segmentation results on a knee CT scan.} 
\label{fig:knee_ct_scan}
\end{figure}

\begin{figure}
\centering
	\begin{tabular}{ccc}
	\rotatebox{90}{\phantom{aaaaaa}Mouse Brain Scan}
	\includegraphics[height=4.8cm, width=4.8cm]{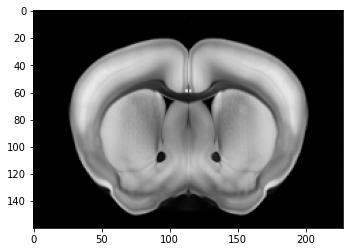}
	&\includegraphics[height=4.8cm, width=4.8cm]{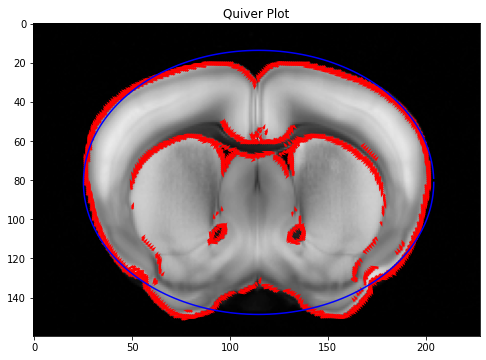}
	&\includegraphics[height=4.8cm, width=4.8cm]{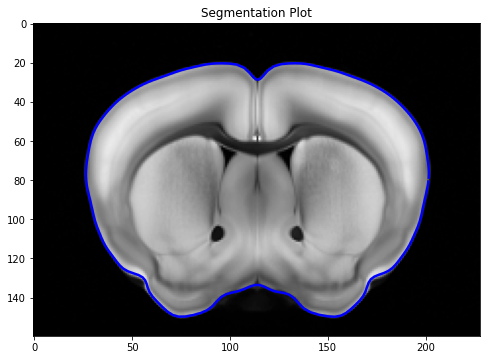}\\
	Source & Gradient Field & Segmentation
	\end{tabular}
\caption{Segmentation results on a mouse brain scan.} 
\label{fig:mouse_brain_scan}
\end{figure}

We next show results of the same procedure applied to two types of medical images: a knee CT scan provided by Dr. Wojtek Zbijewski's lab in Figure \ref{fig:knee_ct_scan} and an MRI mouse brain scan in Figure \ref{fig:mouse_brain_scan}.

\subsection{Results on noisy images}
\begin{figure}
\centering
\begin{tabular}{ccccc}
    \rotatebox{90}{\phantom{aaaaaaa}Bottle Image}
    \rotatebox{90}{\phantom{aaaa}with Gaussian noise}
    &\includegraphics[height=4.65cm, width=4.65cm]{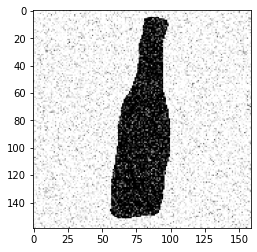}
    &\includegraphics[height=4.65cm, width=4.65cm]{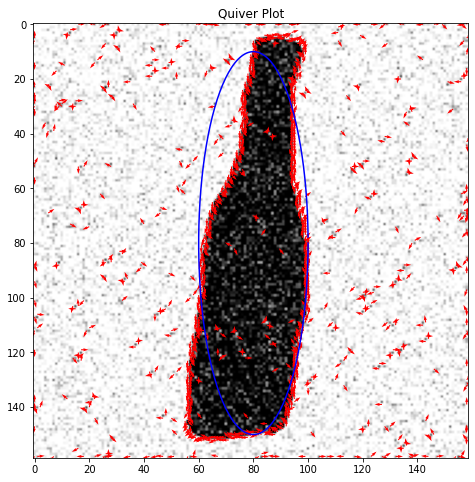}
    &\includegraphics[height=4.65cm, width=4.65cm]{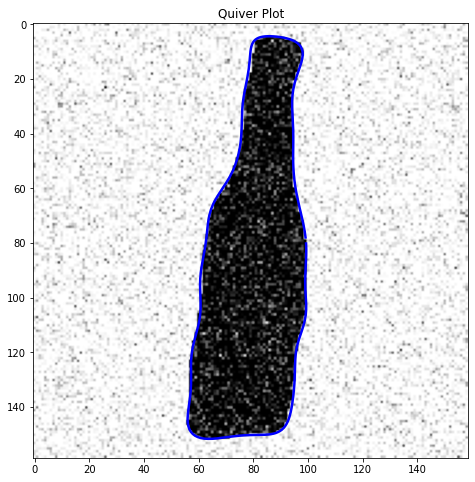}\\
    
    \rotatebox{90}{\phantom{aaaaaaa}Turtle Image}
    \rotatebox{90}{\phantom{aaa}with Salt-and-pepper noise}
    &\includegraphics[height=4.65cm, width=4.65cm]{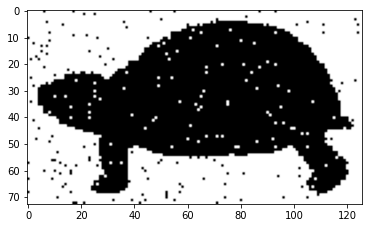}
    &\includegraphics[height=4.65cm, width=4.65cm]{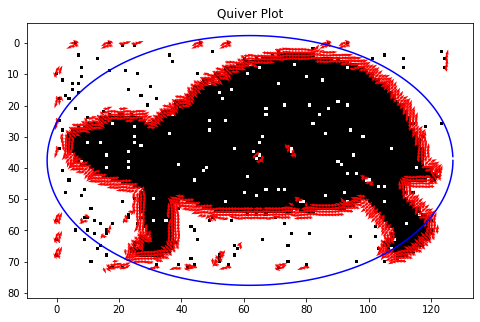}
    &\includegraphics[height=4.65cm, width=4.65cm]{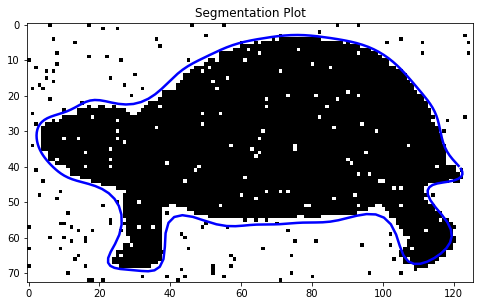}\\
    
    \rotatebox{90}{\phantom{aaaaaaa}Bottle Image}
    \rotatebox{90}{\phantom{aaaa}with Gaussian noise}
    &\includegraphics[height=4.65cm, width=4.65cm]{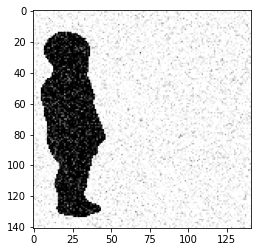}
    &\includegraphics[height=4.65cm, width=4.65cm]{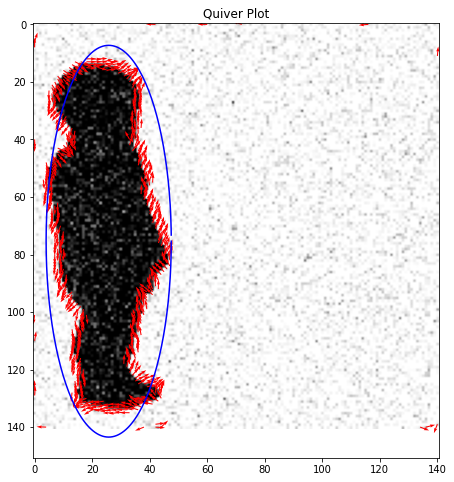}
    &\includegraphics[height=4.65cm, width=4.65cm]{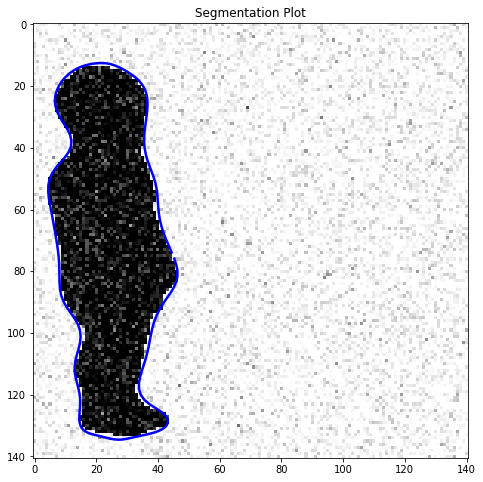}\\
    & Source & Gradient Field & Segmentation 
\end{tabular}	
\caption{Segmentation results on binary images from Kimia database added random noise.} 
\label{fig:binary_image_noise}
\end{figure}

\begin{figure}
\centering
	\begin{tabular}{ccc}
	\rotatebox{90}{\phantom{aaaaaa}Mouse Brain Scan}
	\rotatebox{90}{\phantom{aaa}with Salt-and-pepper noise}
	\includegraphics[height=4.8cm, width=4.8cm]{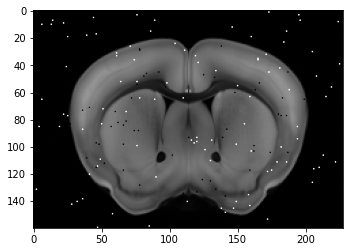}
	&\includegraphics[height=4.8cm, width=4.8cm]{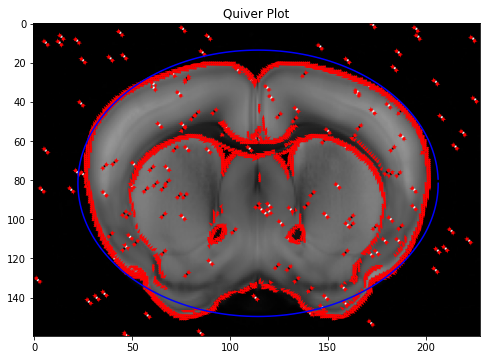}
	&\includegraphics[height=4.8cm, width=4.8cm]{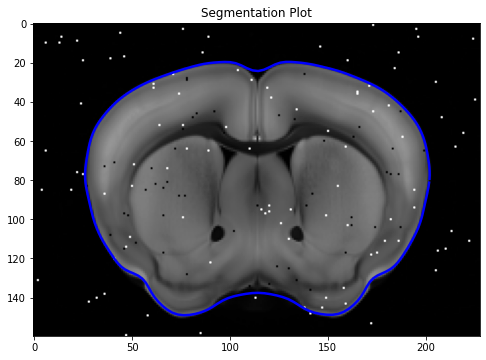}\\
	Source & Gradient Field & Segmentation
	\end{tabular}
\caption{Segmentation results on a mouse brain scan from CLARITY added random noise.} 
\label{fig:mouse_brain_scan_noise}
\end{figure}

Finally, we evaluate the proposed method on images which are corrupted by random noise. Specifically we consider the same images from the Kimia dataset as well as the mouse brain scan of the previous section to which we add either Gaussian or salt and pepper noise. By tuning the threshold for the extraction of the gradients appropriately, we see in Figure \ref{fig:binary_image_noise} and Figure \ref{fig:mouse_brain_scan_noise}  that the segmentation results remain quite good showing that the approach provides some degree of robustness to image noise. 

\section{Conclusion}
   We introduced and implemented a new approach for image segmentation that combined the LDDMM model to represent diffeomorphic transformations of a template curve with a particular loss function to enforce the alignment of this curve to the main gradient directions in the image. As a preliminary proof-of-concept, we illustrated its capability on various types of images including CT and MRI scans as well as images with added random noise. We note that, being a variational approach, this model does not require training over a set of pre-segmented images and thus may be particularly relevant for segmentation of medical images for which only few samples are available.  
   
   However, future work will be needed, first to carry out a rigorous and comprehensive benchmarking with state-of-the-art segmentation methods and then to extend that model to the case of 3D images. Also, one of the main current limitation is the need to select an initial template curve that serves as a prior for the segmentation algorithm. An important next step will be to automatize this process further in particular in the situation of multiple different objects or sub-structures being present in the image.       

\bibliographystyle{unsrt}
\bibliography{reference}

\end{document}